# Vision Transformers and Convolutional Neural Networks for Land Use Scene Classification

Arun D. Kulkarni
Computer Science Department, University of Texas at Tyler, Tyler, TX 75799, USA[1]
Email: akulkarni@uttyler.edu

*Abstract*— **Land Use Scene Classification (LUSC) from remote sensing imagery plays a critical role in environmental monitoring, urban planning, and sustainable resource management. In recent years, deep learning methods have significantly advanced the state of the art, with Convolutional Neural Networks (CNNs) dominating the field because of their strong ability to capture local spatial features. However, the emergence of Vision Transformers (ViTs) has introduced a new paradigm that models long-range dependencies through self-attention mechanisms, potentially enabling improved global context understanding. This paper presents a comparative assessment of Vision Transformers and CNN-based architecture for remote sensing land use scene classification. Representative CNN models, such as AlexNet, is evaluated alongside the Vision Transformer (ViT) using benchmark remote sensing datasets, including the UC Merced (UCM) Land Use and EuroSAT Land Use datasets. The study examines classification accuracy, precision, recall, F1-score, and computational complexity to provide a comprehensive performance comparison. Experimental results demonstrate that CNNs perform robustly on datasets with limited training samples and strong local texture characteristics, whereas Vision Transformers exhibit superior performance in capturing global spatial relationships in complex scenes when sufficient training data are available. However, ViTs typically require greater computational resources and larger training datasets to achieve optimal performance. The findings of this study provide insights into the strengths and limitations of both architectures and offer guidance for selecting appropriate models for remote sensing land use scene classification applications.**

## I. INTRODUCTION

Remote sensing imagery has emerged as a vital data source for analyzing, monitoring, and managing the Earth's surface. Recent advances in satellite and aerial imaging technologies have enabled the acquisition of large volumes of high-resolution imagery, providing unprecedented opportunities for large-scale analysis of land use and land cover patterns [1]. Among the many tasks in remote sensing image analysis, Land Use Scene Classification (LUSC) has emerged as a fundamental problem [2]. LUSC seeks to classify aerial or satellite image patches by assigning semantic labels such as residential, agricultural, industrial, forest, and water bodies. Precise land use scene classification plays a vital role in numerous applications, such as urban planning, environmental monitoring, disaster management, resource allocation, and climate change evaluation.

Early approaches to LUSC primarily used conventional machine learning methods that depended on manually engineered feature extraction followed by statistical classification approaches. These methods typically utilize descriptors such as texture-based features, spectral indices, and shape-based representations. Since local descriptors are usually derived from multiple spatial locations, feature aggregation techniques are needed to generate fixed-length scene representations that can be used for classification tasks. While these methods laid an important foundation for remote sensing image analysis, they also have several intrinsic limitations. In particular, handcrafted features tend to be sensitive to changes in spatial resolution, seasonal variations, lighting conditions, and geographic diversity. Moreover, manually designed descriptors may fail to capture complex semantic relationships among multiple objects and spatial arrangements within a scene [3]. In addition, feature extraction and classifier design are generally treated as separate stages, preventing end-to-end optimization [4]. As a result, traditional methods frequently show poor generalization when used in complex and diverse remote sensing environments.

The rapid expansion of large-scale remote sensing datasets, along with advances in high-performance computing, has driven the increased adoption of deep learning techniques for LUSC. In contrast to traditional machine learning methods, deep neural networks can automatically learn hierarchical feature representations directly from raw image data. This end-to-end representation learning approach has substantially enhanced classification accuracy in remote sensing scenes marked by high intra-class variability, strong inter-class similarity, and considerable spatial heterogeneity.

Among deep learning approaches, Convolutional Neural Networks (CNNs) have emerged as the leading framework for remote sensing image classification. Representative architectures such as AlexNet, VGGNet, ResNet, and EfficientNet have demonstrated strong performance across a variety of benchmark datasets [5]. The success of CNNs is largely attributed to their local receptive fields, parameter sharing, and hierarchical feature extraction mechanisms, which enable efficient learning of edges, textures, shapes, and other discriminative spatial patterns. Moreover, the built-in inductive biases of CNNs—especially locality and translation invariance—tend to encourage more stable learning and improved generalization, particularly in settings with limited training data. Nevertheless, because convolutional operations primarily focus on local neighborhoods, CNNs may be less effective in modeling long-range spatial dependencies and global contextual relationships that frequently characterize complex land use scenes [6].

Recently, Vision Transformers (ViTs) have gained attention as a strong alternative to CNN-based architecture in computer vision. Drawing inspiration from the success of Transformer models in natural language processing, ViTs treat an image as a sequence of non-overlapping patches, which are then processed as tokens. By leveraging self-attention mechanisms, the model directly learns global interactions across different image regions, improving its ability to capture long-range dependencies and contextual relationships. This capacity for global representation is especially valuable in remote sensing imagery, where scene interpretation often relies on spatial relationships among multiple objects spread across wide areas. Despite these advantages, Vision Transformers generally require larger training datasets, higher computational resources, and careful optimization strategies to achieve competitive performance. The differing characteristics of CNNs and Vision Transformers motivate a systematic comparison of their effectiveness in land-use scene classification. CNNs are known for their computational efficiency, strong inductive biases, and reliable performance in data-limited scenarios, whereas Vision Transformers introduce a distinct representation approach that prioritizes global contextual modeling. A rigorous comparative analysis of these architectures is therefore important for understanding their suitability across diverse remote sensing scenarios.

In this paper, we present a comprehensive comparative study of CNN- and ViT-based architectures for land use scene classification using benchmark remote sensing datasets. The study evaluates their classification performance, computational efficiency, and generalization capability under consistent experimental settings. The main objective is to provide practical insights into the strengths and limitations of both model families and to identify conditions under which each architecture is most suitable for remote sensing land use analysis. The remainder of this paper is organized as follows. Section II reviews related work on land-use scene classification. Section III presents the proposed comparative framework and methodology. Section IV describes the datasets, experimental setup, and results. Section V discusses the results, presents the conclusions, and outlines future research directions.

# II. RELATED WORK

Land Use and Land Cover (LULC) scene classification has emerged as a key area of research in remote sensing because of its extensive applications in urban development, environmental surveillance, disaster response, and agricultural evaluation [7, 8]. The growing accessibility of high-resolution satellite imagery has further increased the complexity of scene analysis, creating a demand for more sophisticated computational approaches. Over the last two decades, research in this field has evolved from conventional handcrafted feature extraction techniques to deep learning-based approaches, especially Convolutional Neural Networks (CNNs), and more recently, Vision Transformers (ViTs). This section provides a comprehensive review of these developments and highlights current trends and challenges.

Before the widespread adoption of deep learning techniques, Land Use and Land Cover (LULC) classification primarily relied on handcrafted feature extraction methods integrated with traditional machine learning algorithms. In these conventional approaches, researchers manually engineered descriptive features to capture the spatial, spectral, textural, and structural properties of remote sensing imagery. The effectiveness of the classification process largely depended on the quality and discriminative capability of these handcrafted representations. Among the most influential and extensively utilized feature descriptors were Scale-Invariant Feature Transform (SIFT) [9], Histogram of Oriented Gradients (HOG) [10], and Local Binary Patterns (LBP) [11]. These techniques significantly advanced remote sensing image analysis by enabling the extraction of meaningful visual patterns from satellite and aerial imagery. They were widely employed in combination with classical classifiers such as Support Vector Machines (SVM), Random Forests (RF), k-Nearest Neighbors (k-NN), and Artificial Neural Networks (ANNs) to categorize different land cover and land use classes. SIFT gained considerable attention due to its robustness in detecting distinctive key points that remain invariant to scale, rotation, and partial illumination changes. This characteristic made it highly effective for identifying objects and spatial structures in high-resolution remote sensing images, particularly in heterogeneous environments. HOG descriptors emphasized the distribution of gradient orientations within localized regions, thereby providing strong representations of edge, contour, and shape information. Such capabilities proved valuable for distinguishing complex landscape structures, including urban areas, roads, and vegetation boundaries. Similarly, Local Binary Patterns (LBP) emerged as a powerful texture descriptor because of its simplicity, computational efficiency, and robustness to illumination variations. By encoding local neighborhood intensity relationships, LBP effectively characterized surface textures and fine-grained spatial patterns commonly observed in remote sensing imagery. Consequently, it became widely used in applications involving vegetation analysis, urban mapping, and terrain classification. Collectively, these handcrafted feature extraction techniques established the foundation of traditional LULC classification systems. Although they achieved notable success in many remote sensing applications, their performance was often constrained by limited feature generalization capabilities and the substantial reliance on expert knowledge for feature engineering. These limitations ultimately motivated the transition toward deep learning-based methods, which automatically learn hierarchical and highly discriminative feature representations directly from raw imagery data. Although these methods achieved moderate success, they suffered from several limitations. Most notably, handcrafted features lacked the ability to capture high-level semantic information and were sensitive to variations in scale, illumination, and viewpoint. Furthermore, the feature extraction process required domain expertise and was not adaptable to diverse datasets. These shortcomings motivated the transition toward data-driven feature learning approaches.

The advent of deep learning, particularly Convolutional Neural Networks (CNNs), brought a transformative shift to the field of image classification. Unlike traditional machine learning approaches that rely heavily on handcrafted feature extraction techniques, CNNs are capable of automatically learning hierarchical and discriminative feature representations directly from raw image data. This ability to perform end-to-end learning significantly reduces the dependence on manual feature engineering while improving the robustness and scalability of classification systems [4]. In the domain of remote sensing, the adoption of CNN-based architecture has led to remarkable improvements in scene classification accuracy and efficiency. Early research demonstrated that deep learning models substantially outperform conventional methods by effectively capturing complex spatial patterns, textures, and semantic information present in high-resolution remote sensing imagery [12]. Consequently, CNNs have become a foundational approach for modern remote sensing image analysis and classification tasks. Pre- trained architectures such as AlexNet [13], VGGNet [14], and GoogLeNet [15] were among the first to be adapted for remote sensing applications through transfer learning.

A major strength of CNNs is their ability to learn and capture local spatial patterns through convolutional filters. This capability makes them highly effective for detecting textures, edges, and object-level features in remote sensing imagery [16]. To further improve CNN-based models, researchers have introduced several advancements, including: (a) multi-scale feature extraction techniques to accommodate objects of varying sizes in satellite images [17], and (b) data augmentation and transfer learning strategies to mitigate the challenges posed by limited labeled datasets [18]. Despite these advancements, CNNs have inherent limitations. Their dependence on local receptive fields limits their ability to capture long-range dependencies and global context [6]. Although deeper architectures can partly address this issue, they increase computational complexity and the risk of overfitting, while performance gains often plateau with greater depth. These challenges highlight the need for alternative architecture.

Natural Language Processing-based transformer architectures have significantly influenced the development of Vision Transformer (ViT) models, which have emerged as a powerful alternative to traditional convolutional neural networks (CNNs). Originally inspired by the transformer framework introduced by Vaswani, et al [19]. ViTs depart from the convolution-centric paradigm by treating images as sequences of smaller patches, analogous to tokens in language models. These patches are processed through self-attention mechanisms that enable the model to learn long-range and global dependencies across the entire image more effectively than conventional CNN architectures. The pioneering study by Dosovitskiy et al. [6] demonstrated that ViTs are capable of achieving state-of-the-art performance in image classification tasks, particularly when trained on large-scale datasets. Their findings highlighted the scalability and representational power of transformer-based vision models, establishing ViTs as a transformative approach in computer vision research. In the domain of remote sensing, ViTs have shown considerable potential for handling the complex spatial structures and contextual variations commonly present in satellite imagery. Their ability to model global contextual relationships makes them especially suitable for Land Use and Land Cover Classification (LULC) classification tasks, where understanding both local features and broader spatial patterns is essential for accurate categorization. Recent studies, such as those by Zhao et al. [8], have demonstrated that transformer-based approaches can effectively capture intricate spatial dependencies and improve classification performance in remote sensing applications. The global attention mechanism in ViTs enables them to model long-range interactions more effectively than CNNs. This is particularly beneficial for complex scenes where spatial relationships between distant regions are important. However, ViTs also present challenges, including high computational requirements, large memory consumption, and dependence on extensive training data [20]. To harness the complementary advantages of CNNs and transformers, recent studies have increasingly explored hybrid architectures that integrate convolutional operations with self-attention mechanisms. These approaches seek to combine the strong local feature extraction capabilities of CNNs with the powerful global context modeling offered by transformers. Research indicates that hybrid models frequently outperform standalone CNNs and Vision Transformers (ViTs) in Land Use and Land Cover (LULC) classification tasks [21]. They demonstrate greater robustness, improved generalization, and a superior ability to capture multi-scale contextual information. Such models are especially well suited for analyzing high-resolution and hyperspectral remote sensing datasets. The progress of Land Use and Land Cover (LULC) classification techniques has been strongly driven by the availability of publicly accessible benchmark datasets that enable standardized model training and fair performance comparison across studies. Datasets in this domain include the UCM Land Use Dataset and the EuroSAT Dataset.

The UCM dataset provides high-resolution aerial imagery with a limited number of well-balanced classes, whereas EuroSAT, derived from Sentinel-2 satellite imagery, further extends this diversity by incorporating multispectral information. These datasets form a comprehensive benchmark suite that enables robust evaluation of model generalization, feature representation capability, and cross-dataset transferability [17, 22]. Convolutional Neural Network (CNN)-based architectures have historically demonstrated strong performance across these benchmarks due to their effectiveness in learning hierarchical spatial features from imagery. However, recent advances in deep learning have shown that transformer-based models, such as Vision Transformers (ViTs), along with hybrid CNN–Transformer architectures, are achieving either superior or competitive performance. These models are particularly effective in capturing long-range spatial dependencies and global contextual relationships, which are often challenging for traditional CNNs. As a result, there is a growing shift toward attention-based and hybrid frameworks in LULC classification, especially for complex and large-scale remote sensing datasets. Duan et al. [23] proposed the Swin Transformer with Multi-Scale Fusion (STMSF) to effectively represent ground objects of varying scales in remote sensing scenes through the integration of multi-scale features.

Experimental results show that the proposed network surpasses several state-of-the-art CNN-based and transformer-based approaches.

Recent studies in remote sensing and land use/land cover (LULC) classification demonstrate a clear and growing transition toward transformer-based architectures and hybrid deep learning models. These approaches have shown strong potential in capturing long-range spatial dependencies and improving classification performance compared to traditional convolutional methods. Despite these advancements, several important challenges persist. First, data scarcity remains a major limitation, as high-quality, accurately labeled remote sensing datasets are often limited in size and diversity, constraining the training of data-intensive models. Second, computational complexity is a significant concern, particularly for transformer-based models, which typically require substantial memory and processing power for both training and inference. Third, model interpretability continues to be an open issue, as the internal decision-making processes of deep learning models—especially transformers—are often difficult to interpret and explain in a physically meaningful way. Finally, generalization across datasets remains challenging, since models trained on a specific dataset often experience performance degradation when applied to different geographic regions, sensor types, or imaging conditions.

In this context, the present study focuses on a comparative analysis of Convolutional Neural Networks (CNNs) and Vision Transformers (ViTs) for Land Use Scene Classification (LUSC) classification. The evaluation is conducted using two widely recognized benchmark datasets: the UCM Land Use Dataset and the EuroSAT Dataset. This comparative investigation aims to assess the relative strengths and limitations of CNN-based and transformer-based approaches in terms of classification accuracy, robustness, and generalization capability across diverse remote sensing scenarios.

## III. PROPOSED FRAMEWORK

This section presents a comparative framework for evaluating Convolutional Neural Networks (CNNs) and Vision Transformers (ViTs) for land use scene classification (LUSC) using high-resolution remote sensing imagery. To ensure a controlled and fair comparison, both the AlexNet and Vision Transformer (ViT) models are trained using identical datasets and data partitions, input resolutions, augmentation strategies, optimization settings, and evaluation metrics. Therefore, the observed performance differences can be attributed primarily to architectural characteristics rather than experimental inconsistencies. The proposed framework consists of four. stages (a) dataset preparation and preprocessing, (b) AlexNet optimization and training, (c) Vision Transformer (ViT) optimization and training, (d) classification of test data.

### A. Data Preparation and Preprocessing

Let the land use dataset be represented as

$$\mathcal{D} = \{(x_i, y_i)\}_{i=1}^{N} \quad (1)$$

Where $x_i \in \mathbb{R}^{H \times W \times C}$ , denotes an input image, $y_i \in \{1,2, \dots, K\}$ is the class label, $N$ is the total number of samples, and $K$ is the number of land use categories. Remote sensing images exhibit substantial intra-class variability arising from seasonal effects, spatial resolution changes, illumination differences, and scene complexity. To reduce statistical variability across samples, each image is normalized using channel-wise standardization:

$$x_i' = \frac{x_i - \mu}{\sigma} \quad (2)$$

where μ and σ denote channel mean and standard deviation. To improve generalization and reduce overfitting, stochastic augmentation is applied during training:

$$\tilde{x}_i = T(x_i') \quad (3)$$

where $T(\cdot)$ denotes a composition of random transformations including horizontal flipping, random rotation, random cropping, and mild color perturbation. The dataset is partitioned into mutually exclusive subsets:

$$\mathcal{D} = \mathcal{D}_{train} \cup \mathcal{D}_{val} \cup \mathcal{D}_{test} \quad (4)$$

With

$$\mathcal{D}_{train} \cap \mathcal{D}_{val} = \emptyset, \mathcal{D}_{train} \cap \mathcal{D}_{test} = \emptyset \quad (5)$$

### B. AlexNet for Land Use Classification

CNNs exploit local spatial correlations through learnable convolutional kernels. In this study, AlexNet was implemented using the MATLAB Deep Learning Toolbox. The architecture of AlexNET is shown in Fig. 1. The model consists of eight layers, including convolutional layers, max-pooling layers, fully connected layers, and a SoftMax output layer. The functions of the AlexNet layers are explained using the equations presented below [13, 24].

Given the input feature map $X^{(l-1)}$ , the convolutional transformation at layer l is defined as

$$X_j^{(l)} = \phi\left(\sum_{i=1}^{M} X_i^{(l-1)} * W_{ij}^{(l)} + b_j^{(l)}\right) \quad (6)$$

Where $W_{ij}^{(l)}$ denotes learnable convolution kernels, $b_j^{(l)}$ is bias, and ϕ(·) is the nonlinear activation function.

Pooling performs spatial down sampling:

$$P^{(l)} = \text{Pool}\left(X^{(l)}\right) \quad (7)$$

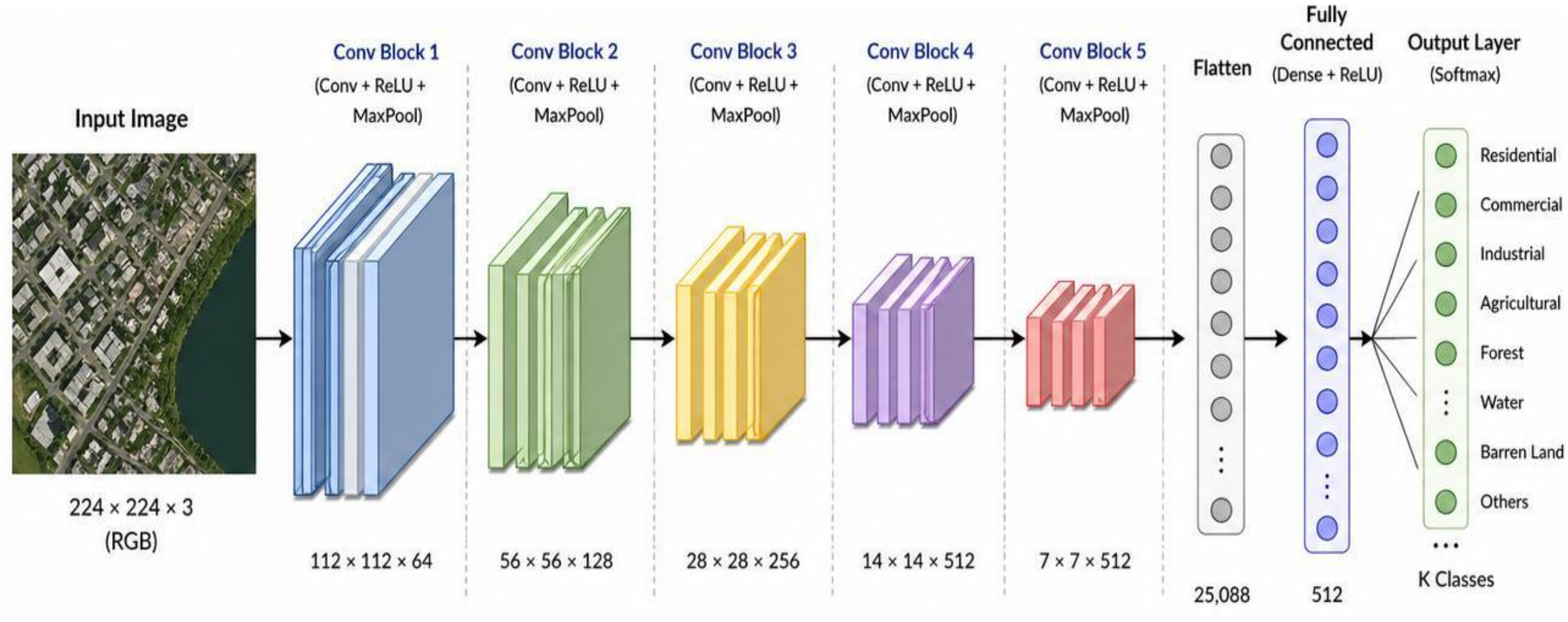

Fig. 1. AlexNet Architecture for Land Use Classification

The hierarchical nature of convolution allows the network to learn low-level edge patterns in early layers and increasingly abstract semantic structures in deeper layers. After feature extraction, the resulting representation is flattened and passed through fully connected layers. The final class probability is obtained using SoftMax:

$$\hat{y}_k = \frac{\exp(z_k)}{\sum_{j=1}^{K} \exp(z_j)} \quad (8)$$

Where $z_k$ denotes the class logit. The SoftMax classifier converts logits into normalized probability distributions for multi-class prediction tasks [25]. CNNs are expected to perform strongly when land use categories are characterized by repetitive textures, locally correlated spatial patterns, and compact object arrangements. Examples include agricultural land, forest regions, and dense residential structures [26].

### C. Vision Transformer for Land Use Classification

A Vision Transformer (ViT) is a neural network architecture for computer vision that applies the Transformer model, originally developed for natural language processing, to image analysis. It operates by dividing an image into small patches, converting each patch into an embedding vector, adding positional information, and processing the resulting sequence of embeddings through self-attention layers to learn relationships among different regions of the image. Unlike traditional Convolutional Neural Networks (CNNs), which primarily focus on local feature extraction using convolutional filters, ViTs can capture long-range global dependencies more effectively. This capability makes them highly effective for tasks such as image classification, object detection, and image segmentation, particularly when trained on large-scale datasets. A simplified architecture of the Vision Transformer is shown in Fig. 2. The ViT encoding mechanism is explained below with the help of the corresponding equations [6, 19, 27].

#### *1) Patch Embedding*

Given an input image

$$x \in \mathbb{R}^{H \times W \times C} \quad (9)$$

It is divided into non-overlapping patches of size P×P. The number of patches is

$$N_p = \frac{HW}{P^2} \quad (10)$$

Each patch is flattened and projected into a D-dimensional embedding space:

$$z_0^p = x_p E \quad (11)$$

where $E \in \mathbb{R}^{(P^2 C) \times D}$ is a learnable projection matrix.

A learnable class token is appended and positional embeddings are added:

$$Z_0 = \left[x_{cls}; z_0^1; z_0^2; \ldots; z_0^{N_p}\right] + E_{pos} \quad (12)$$

#### *2) Multiheaded Attention*

For an input representation Z, query, key, and value matrices are defined as

$$Q = ZW_Q, \quad K = ZW_K, \quad V = ZW_V \quad (13)$$

Self-attention is computed as

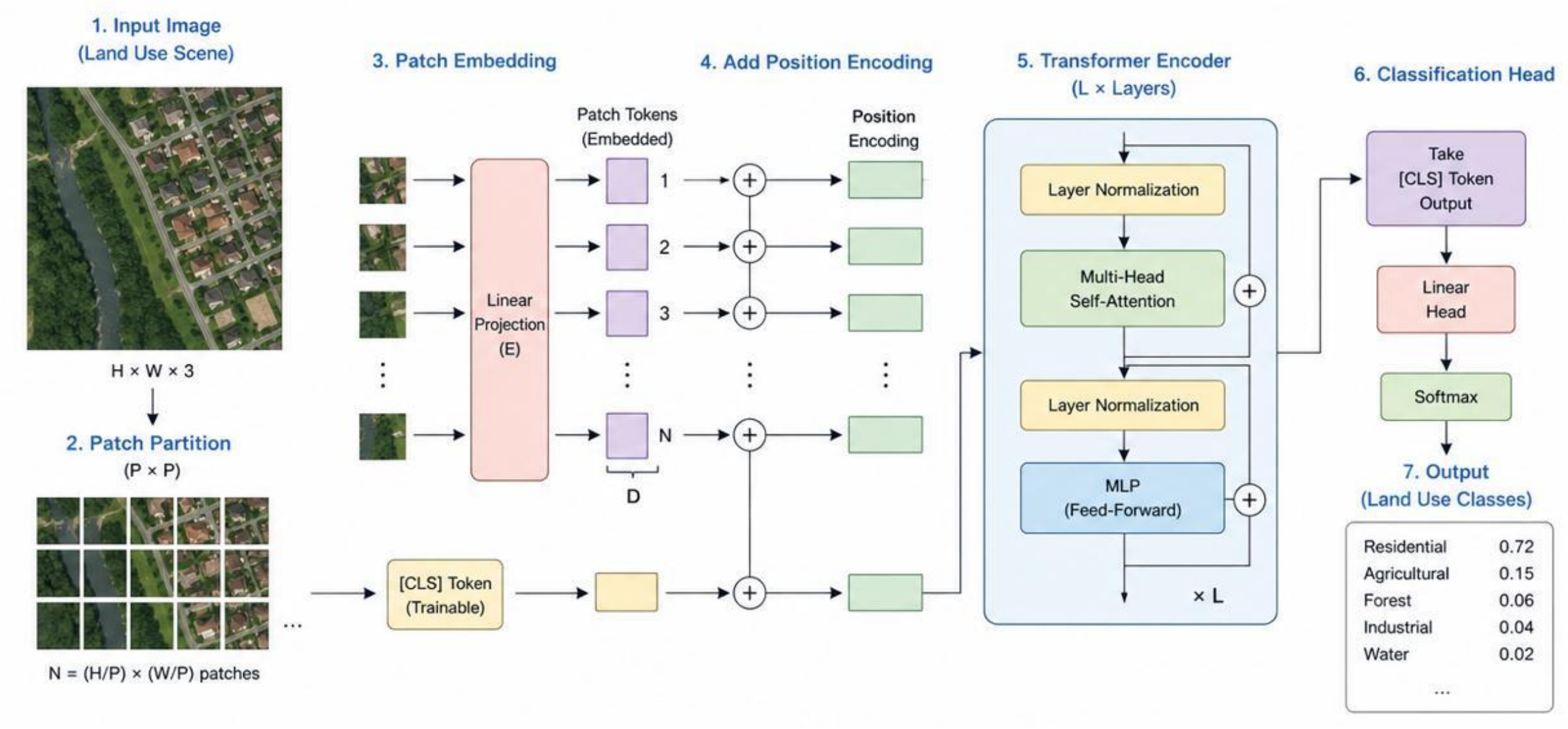

Fig. 2. Vision Transformer Architecture for Land Use Scene Classification

$$\text{Attention}(Q, K, V) = \text{Softmax}\left(\frac{QK^T}{\sqrt{d_k}}\right)V \quad (14)$$

Multi-head attention is expressed as

$$\text{MSA}(Z) = \text{Concat}(head_1, \ldots, head_h)W_O \quad (15)$$

where $h$ denotes the number of attention heads.

*3) Transformer Encoding*

Each encoder layer contains residual self-attention and feed-forward blocks:

$$Z_l' = Z_{l-1} + \text{MSA}\big(\text{LN}(Z_{l-1})\big) \quad (16)$$

$$Z_l = Z_l' + \text{MLP}\big(\text{LN}(Z_l')\big) \quad (17)$$

The final class token is used for classification:

$$\hat{y} = \text{Softmax}(W_c Z_L^{cls} + b_c) \quad (18)$$

ViTs are particularly suitable when scene understanding depends on global spatial organization and long-range contextual relationships. Examples include airports, harbors, commercial zones, and transportation hubs.

*D. Optimization Strategy*

Both models are trained under identical optimization settings. The training objective is categorical cross-entropy.

$$\mathcal{L} = -\frac{1}{N}\sum_{i=1}^{N}\sum_{k=1}^{K} y_{ik}\log(\hat{y}_{ik}) \quad (19)$$

Parameter updates are performed by gradient-based optimization:

$$\theta_{t+1} = \theta_t - \eta\nabla_\theta\mathcal{L} \quad (20)$$

where θ denotes trainable parameters and η is the learning rate [28].

To improve generalization, the framework incorporates early stopping, weight decay, dropout regularization, and learning-rate scheduling.

*E. Evaluation Protocol*

Performance is evaluated using multiple complementary metrics [25].

$$\text{Accuracy} = \frac{TP + TN}{TP + TN + FP + FN} \quad (21)$$

$$\text{Precision} = \frac{TP}{TP + FP} \quad (22)$$

$$\text{Recall} = \frac{TP}{TP + FN} \quad (23)$$

$$F_1 = \frac{2PR}{P + R} \quad (24)$$

Confusion matrices are used to analyze class-level error patterns and identify categories that benefit from local or global feature modeling. The proposed approach establishes a rigorous experimental framework for comparing convolutional and transformer-based models for land use scene classification. By enforcing identical preprocessing, training, and evaluation conditions, the framework enables an interpretable assessment of architectural behavior. The next section describes experimental setup and results.

## IV. IMPLEMENTATION AND RESULTS

This section describes the datasets, implementation details, and protocols used to assess the effectiveness of the AlexNet and Vision Transformer framework for land use scene classification (LUSC).

### *A. Datasets*

To evaluate the robustness and generalization capability of the CNN and ViT models, experiments were conducted on two widely used benchmark remote sensing datasets: the UCM Land Use Dataset and the EuroSAT Dataset. For both models and across all datasets, 80 percent of the randomly selected samples from each class were used for training, 10 percent were used for validation, and the remaining 10 percent were used for testing.

The UCM Land Use Dataset is a widely used benchmark dataset in the field of remote sensing and aerial image classification. Developed by researchers at the University of California, Merced [29]. The dataset contains 2,100 high-resolution aerial images divided into 21 land-use categories, including agricultural, forest, harbor, residential, runway, and parking lot scenes. Each category contains 100 images with image dimensions of 256 × 256 pixels and a spatial resolution of 0.3 meters (approximately one foot per pixel). The dataset is commonly used to evaluate machine learning and deep learning models for scene classification, feature extraction, and pattern recognition in geospatial imagery. Owing to its balanced class distribution and diverse visual content, the dataset has become a standard benchmark for comparing the performance of convolutional neural networks and other computer vision approaches in remote sensing research. For this study, two subsets were created from the UCM Land Use Dataset. The first subset consists of five distinct classes, while the second subset contains ten classes, including the original five classes and five additional classes. Images from both subsets were classified using the AlexNet and Vision Transformer models, implemented using MATLAB scripts. Forty sample images from UCM Land Use 10 class Data set are shown in Fig. 3.

EuroSAT Dataset is a widely used benchmark dataset for land-use and land-cover classification in remote sensing applications [30]. The dataset is based on satellite images acquired from the Sentinel-2 mission of the European Space Agency. It contains 27,000 labeled and geo-referenced images categorized into 10 different land-use and land-cover classes, including annual crops, forest, herbaceous vegetation, highway, industrial, pasture, permanent crop, residential, river, and sea/lake scenes. Each image has dimensions of 64 × 64 pixels and covers a spatial resolution ranging from 10 to 60 meters per pixel, depending on the spectral band used. The dataset is extensively used for evaluating machine learning and deep learning algorithms in remote sensing image classification, scene understanding, and environmental monitoring tasks. Due to its large number of samples, balanced class distribution, and multispectral image characteristics, the EuroSAT Dataset has become a standard benchmark for comparing the performance of convolutional neural networks, transformer-based architectures, and other computer vision models in satellite image analysis. In this study, we created two sets of data the first data set consists of 5 distinct classes, and the second data set consists of all ten classes from the EuroSAT Dataset. Sample images from EuroSAT data set with 10 classes are shown on Fig. 4.

Fig. 3. Sample images from UCM Dataset

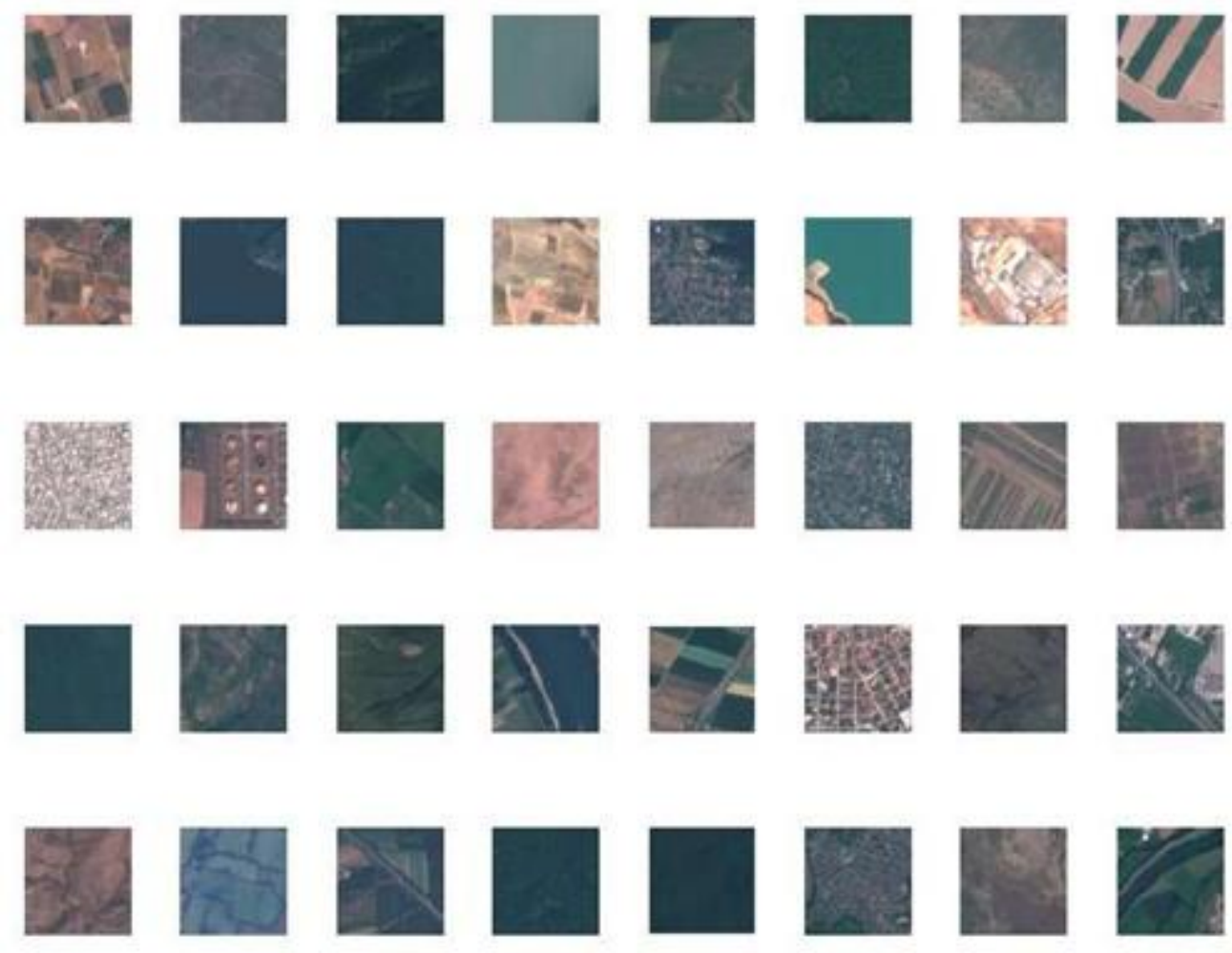

Fig. 4. Sample Images from EuroSAT Dataset

All datasets were classified using the AlexNet and Vision Transformer models implemented in MATLAB on a PC equipped with a single AMD Ryzen 9 7950X x64 4.5 GHz processor and 128 GB RAM. Evaluation metrics, including confusion matrices, accuracy, precision, recall, F-score, and computational time, are presented in this section.

### *B. Network Configuration*

We developed two AlexNet models using MATLAB scripts. Each model consists of eight learnable layers, including five convolutional layers and three fully connected layers. The input image size was 227 × 227 × 3 (RGB). The first model was designed for five output classes, while the second model was designed for ten output classes. The layer-by-layer details are presented in Table 1. In addition, we developed two Vision Transformer B/16 models using MATLAB scripts. The input image size was 224 × 224 × 3 (RGB), with a patch size of 16 × 16 pixels, resulting in 196 patches. The first ViT model was designed for five classes, while the second model was designed for ten classes. The feature details of the Vision Transformer are presented in Table 2. The experimental setup ensures a fair, rigorous, and reproducible evaluation of the CNN and Vision Transformer models across multiple benchmark datasets. By incorporating strong baselines, evaluation metrics described in Section III. The framework provides a comprehensive assessment of model performance.

TABLE 1. ALEXNET LAYER-BY-LAYER DETAILS

| Layer | Type | Filters / Neurons | Kernel | Stride | Output Size |
|---|---|---|---|---|---|
| Input | Image | — | — | — | 227×227×3 |
| Conv1 | Convolution | 96 filters | 11×11 | 4 | 55×55×96 |
| Pool1 | Max Pool | — | 3×3 | 2 | 27×27×96 |
| Conv2 | Convolution | 256 filters | 5×5 | 1 | 27×27×256 |
| Pool2 | Max Pool | — | 3×3 | 2 | 13×13×256 |
| Conv3 | Convolution | 384 filters | 3×3 | 1 | 13×13×384 |
| Conv4 | Convolution | 384 filters | 3×3 | 1 | 13×13×384 |
| Conv5 | Convolution | 256 filters | 3×3 | 1 | 13×13×256 |
| Pool5 | Max Pool | — | 3×3 | 2 | 6×6×256 |
| FC6 | Fully Connected | 4096 | — | — | 4096 |
| FC7 | Fully Connected | 4096 | — | — | 4096 |
| FC8 | Fully Connected | 1000 | — | — | 1000 |

### C. Experimental Results

We implemented two AlexNet models and two Vision Transformer (ViT) models to classify UCM and EuroSAT datasets with five and ten classes respectively. The results are presented below

TABLE II. VISION TRANSFORMER-B/16 FEATURES

| Feature | Details |
|---|---|
| Name | ViT-B/16 |
| Patch size | 16×16 pixels |
| Input image | 224×224×3 (standard) |
| Embedding dimension | 768 |
| Number of transformer blocks | 12 |
| Number of attention heads | 12 |
| MLP (Feed-forward) hidden size | 3072 |
| Total parameters | ~86 million |
| Classification token | Yes ([CLS]) |
| Positional encoding | Learnable |

### *1. Results for UCM Dataset*

The learning progress graph for UCM dataset for AlexNet and Vision Transformer are shown in Fig. 5 and Fig. 6, respectively

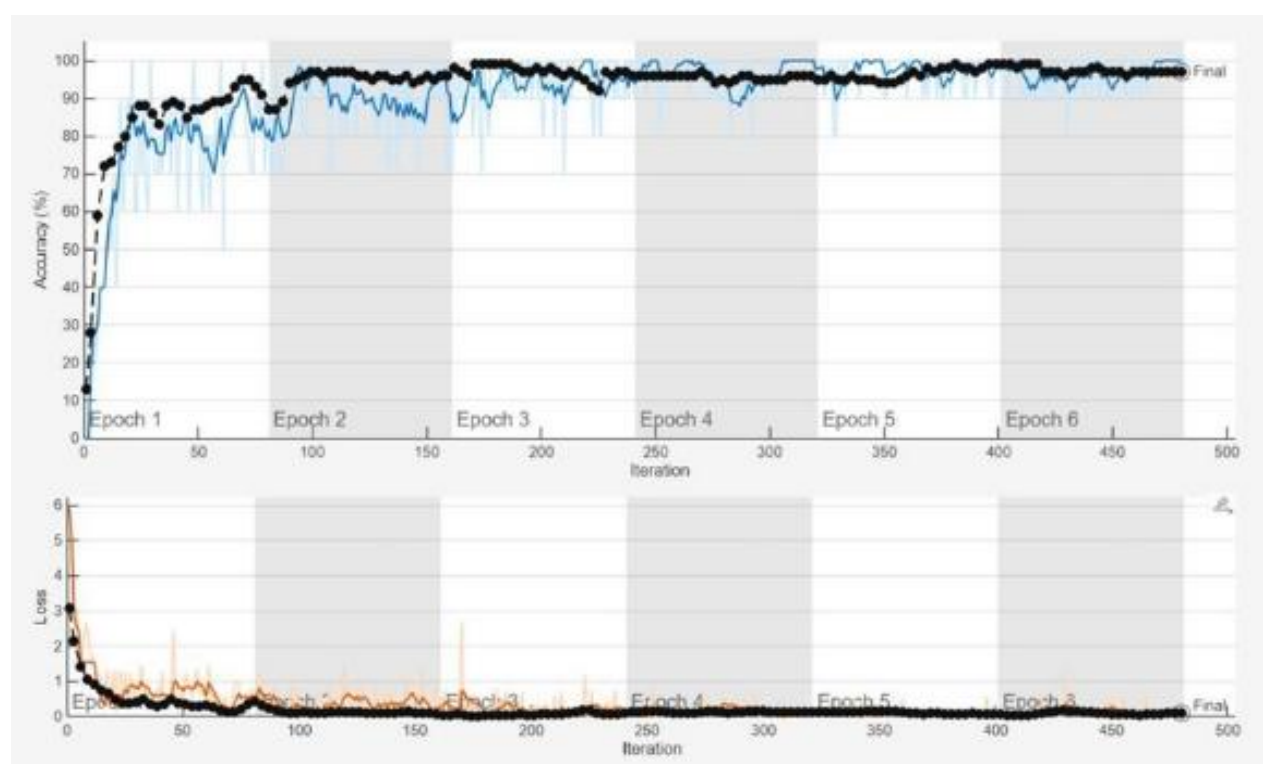

Fig. 5. Learning Progress for AlexNet 10-Class UCM Dataset

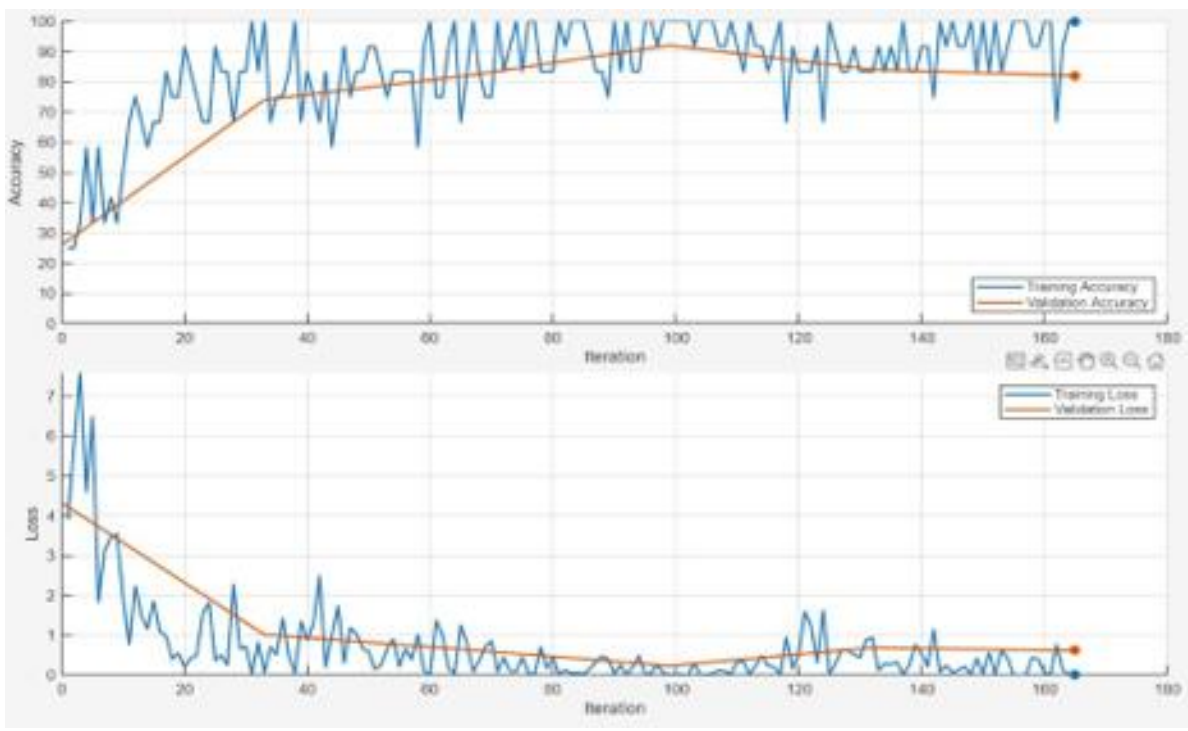

Fig. 6. Learning Progress for ViT 10-Class UCM Dataset

The comparison of the average values of the metrics—accuracy, precision, recall, and F-score—obtained from the confusion matrices for the UCM Dataset using AlexNet and Vision Transformer is shown in Fig. 7 and Fig. 8, respectively.

Sample classified output images with their labels obtained from AlexNet and the ViT are shown in Fig. 9 and Fig. 10.

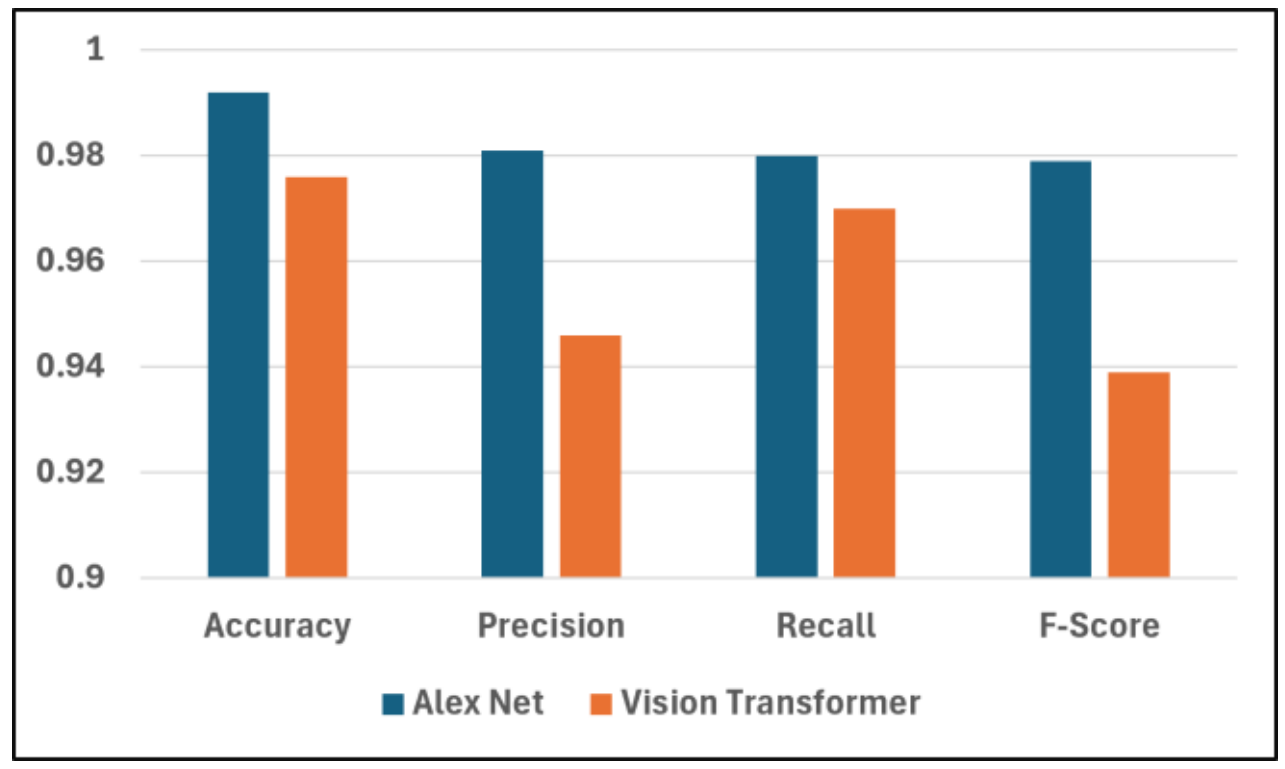

Fig. 7. Comparative Performance for 5-Class UCM Dataset

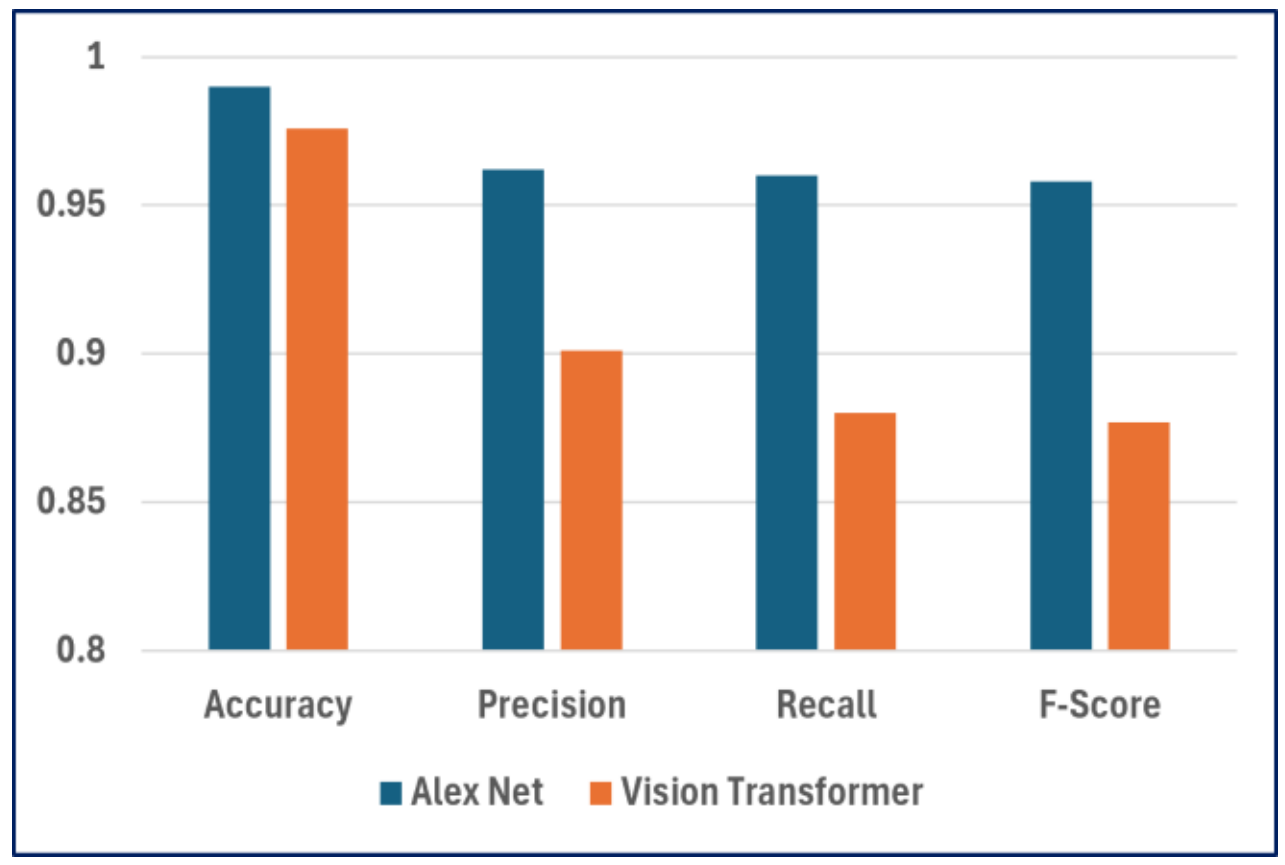

Fig. 8. Comparative Performance for 10-Class UCM Dataset

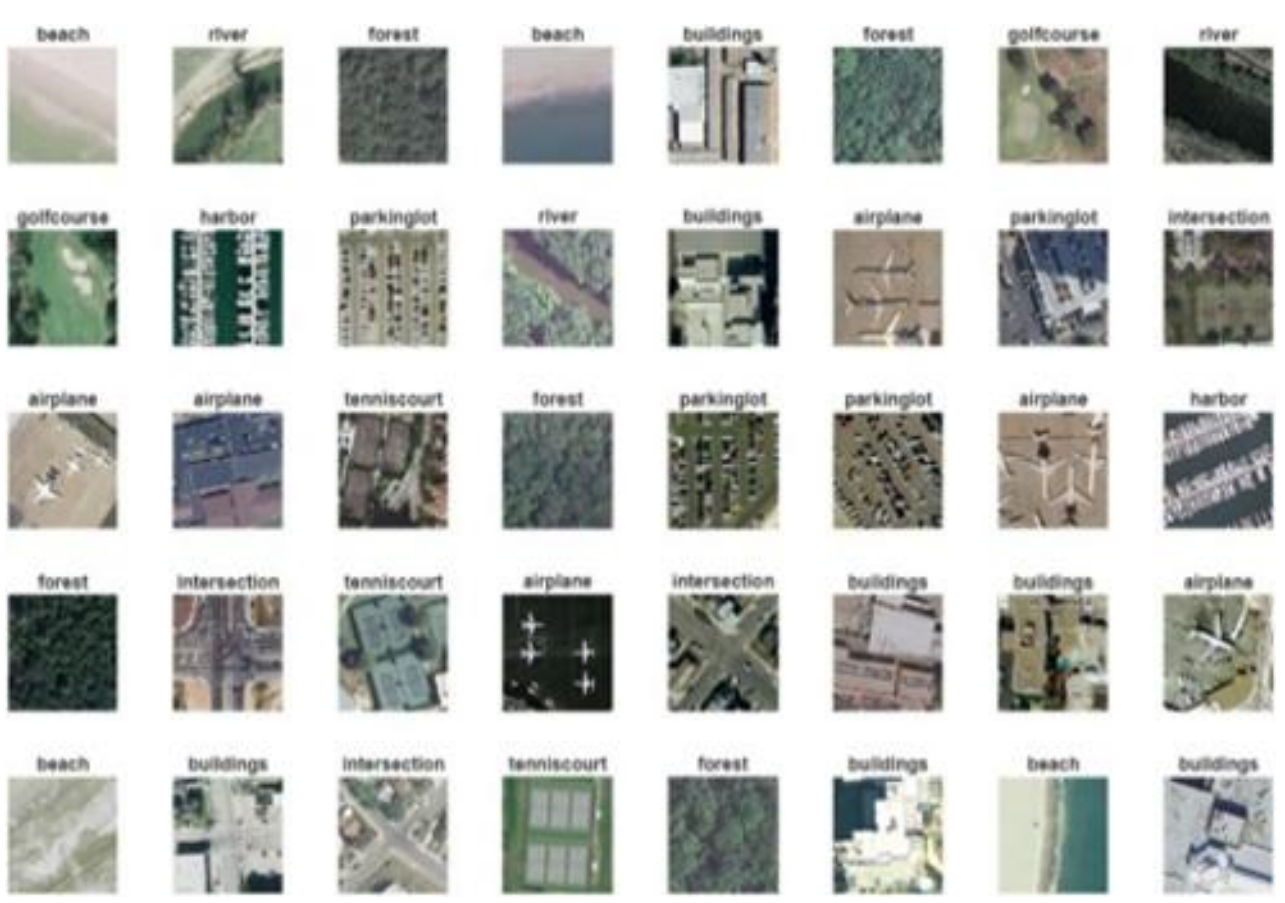

Fig. 9. Classified Output Images with AlexNet

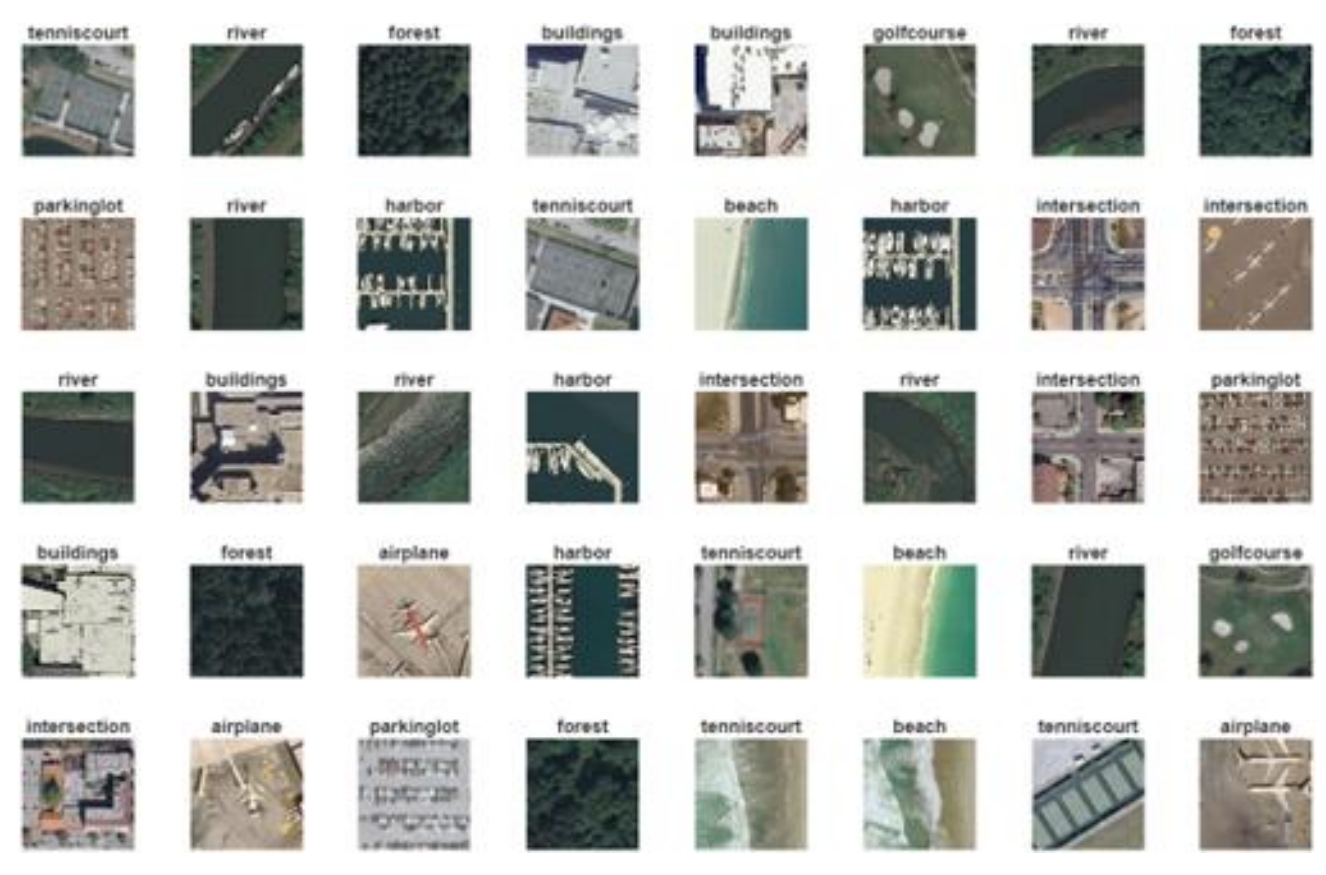

Fig. 10. Classified Output Images With ViT

## 2. Results for EuroSAT Dataset

The learning progress graph for EuroSAT dataset for AlexNet and Vision Transformer are shown in Fig. 11 and Fig. 12, respectively.

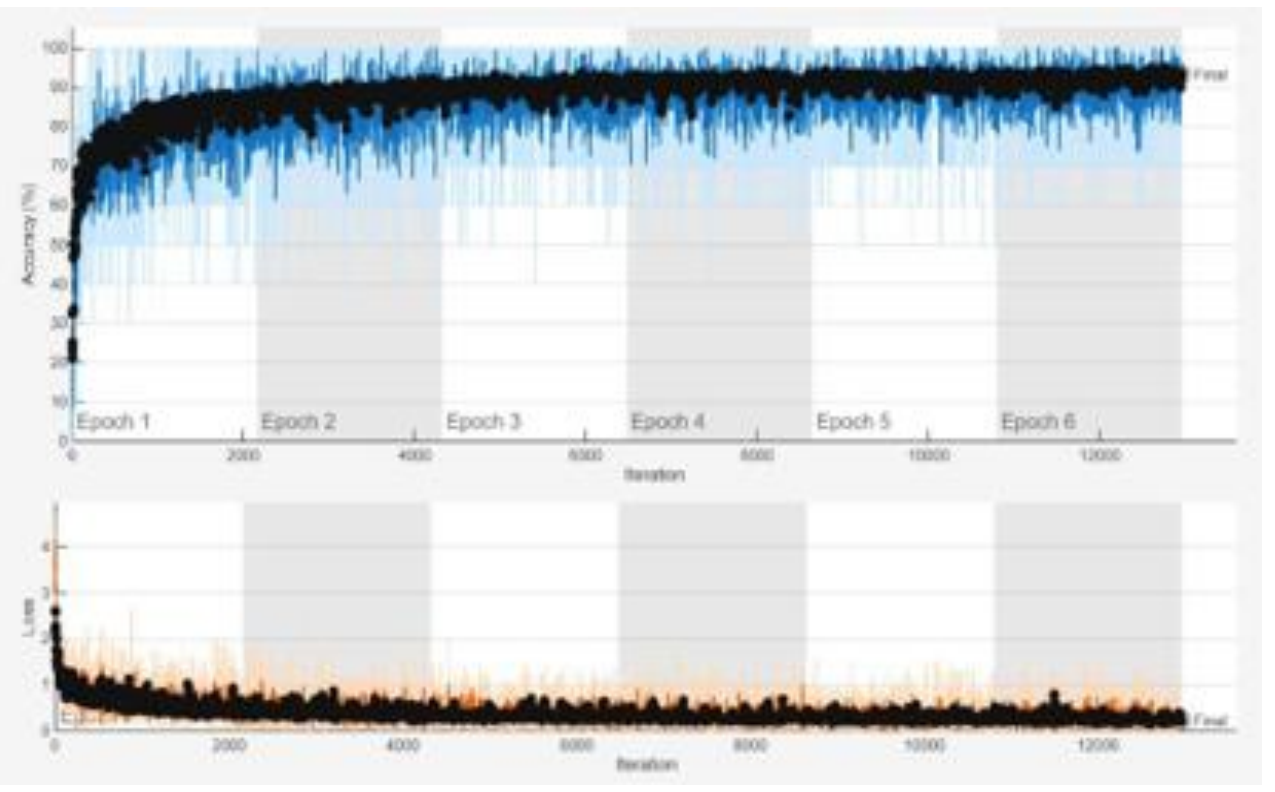

Fig. 11. Learning Progress with AlexNet 10-Class EuroSAT Dataset

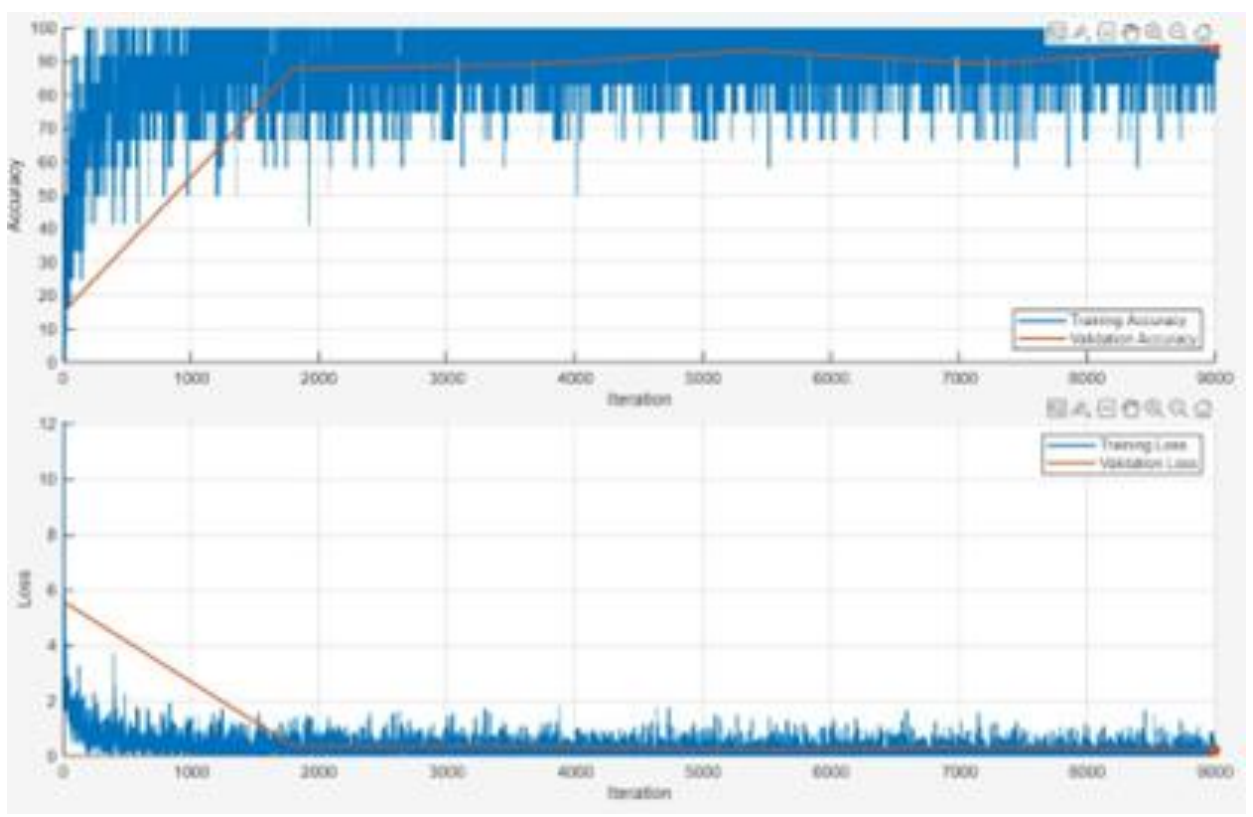

Fig. 12. Learning Progress with ViT 10-Class EuroSAT Dataset

The comparison of the average values of the metrics—accuracy, precision, recall, and F-score—obtained from the confusion matrices for the EuroSAT Dataset using AlexNet and Vision Transformer is shown in Fig. 13 and Fig. 14, respectively. Sample classified output images with their labels obtained from AlexNet and the ViT are shown in Fig. 15 and Fig. 16.

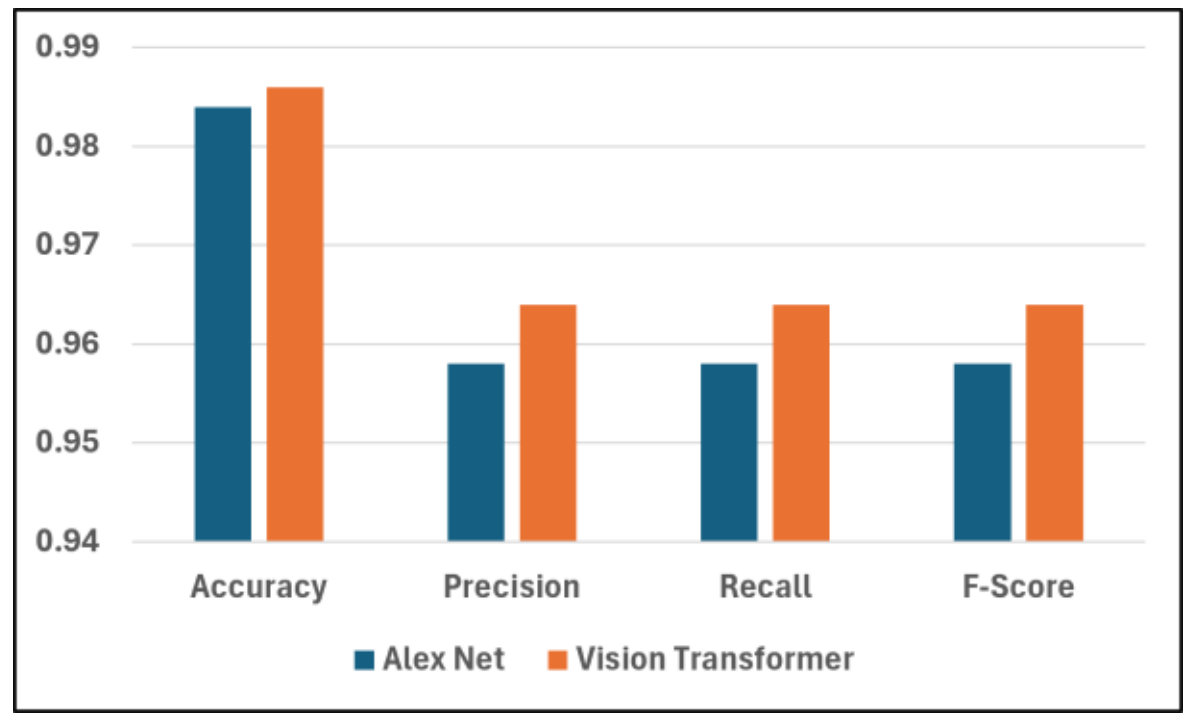

Fig. 13. Comparative Performance for 5-Class EuroSAT Dataset

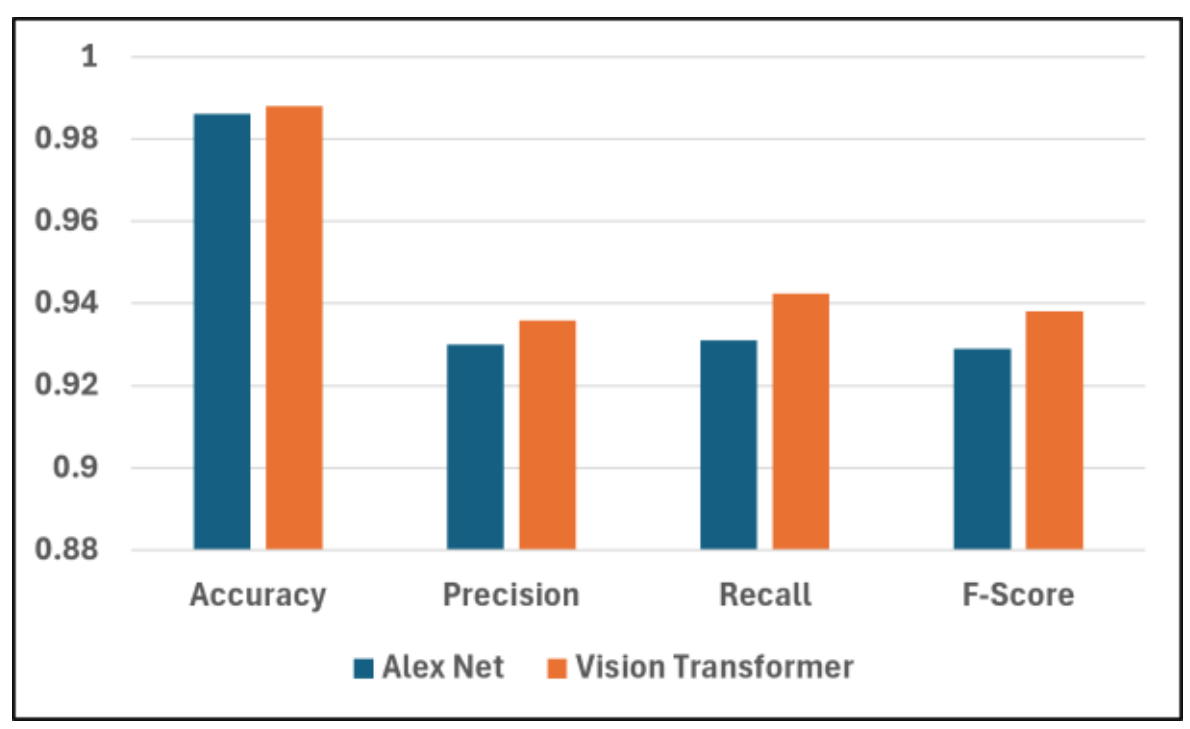

Fig. 14. Comparative Performance for 10-Class EuroSAT Dataset

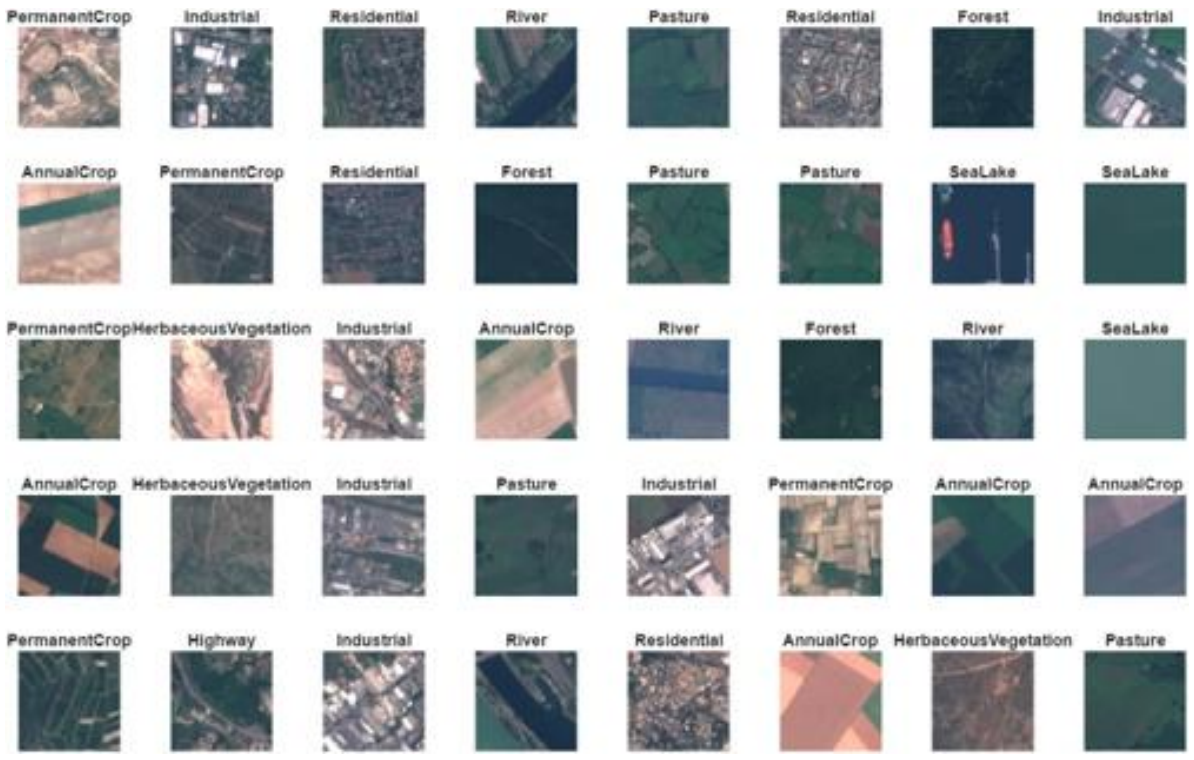

Fig. 15. Classified Output with AlexNet for 10-Class EuroSAT Dataset

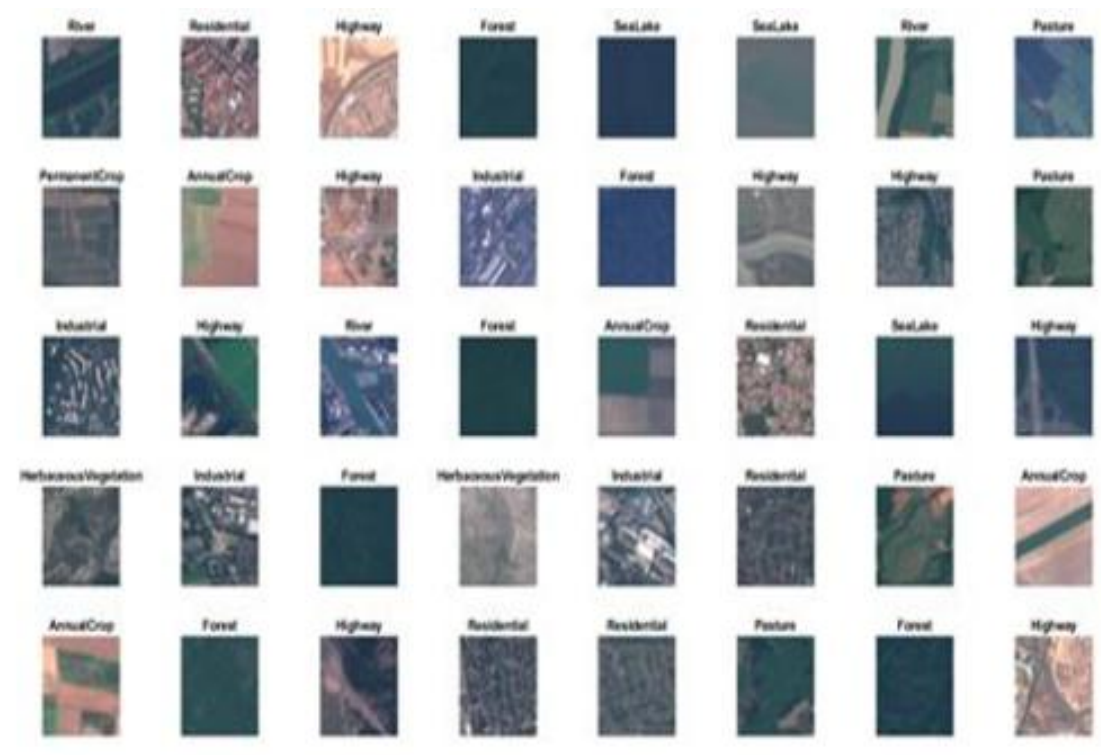

Fig. 16. Classified Output with ViT for 10-Class EuroSAT Dataset

## V CONCLUSION

In this study, we implemented and evaluated the convolutional neural network (CNN)-based AlexNet model and the Vision Transformer (ViT-B16) model for land-use scene classification using benchmark remote sensing image datasets. The experiments were conducted on the 5-class and 10-class versions of the UCM Land Use Dataset and the EuroSAT RGB Dataset. Both models were trained and tested on a system equipped with an AMD Ryzen 9 7950X processor and 128 GB RAM. For both models and all datasets, 80 percent of the randomly selected samples from each class were used for training, 10 percent for validation, and the remaining 10 percent for testing. By enforcing identical preprocessing, training, and evaluation conditions, the proposed framework enables a clear and interpretable assessment of architectural behavior. Model performance was evaluated using standard classification metrics, including accuracy, precision, recall, and F1-score. The evaluation results for the UCM datasets are presented in Figs. 7 and 8. The results indicate that AlexNet achieves superior performance when the dataset contains a relatively small number of training samples. The UCM dataset includes approximately 100 samples per class, which appears to favor CNN-based feature extraction and classification. The evaluation results for the EuroSAT datasets are presented in Figs. 13 and 14. In contrast to the UCM dataset results, the ViT-B16 model demonstrates slightly better performance than AlexNet on the EuroSAT dataset, where the number of samples per class ranges from 2,000 to 3,000 images. This observation suggests that Vision Transformer architectures are more effective when trained on larger-scale datasets, as they can better exploit abundant training data to learn global contextual representations. Despite their improved performance on larger datasets, Vision Transformer models require greater computational resources. In our experiments, training the ViT-B16 model on the 10-class EuroSAT dataset required approximately 253 minutes, whereas AlexNet required only 137 minutes under the same hardware configuration. In addition, ViT models demand greater memory capacity and computational complexity compared with conventional CNN architecture. Overall, the experimental results indicate that CNN-based models such as AlexNet are more suitable for land-use scene classification tasks involving limited training samples, whereas Vision Transformer models achieve better performance when sufficient training data samples are available. However, the improved performance of Vision Transformers comes at the cost of increased computational requirements.

Future research may focus on hybrid architectures that combine the local feature extraction capability of CNNs with the global attention mechanisms of Vision Transformers to further improve land-use scene classification accuracy and robustness. In addition, future work will include the evaluation of other CNN architectures, such as VGGNet, ResNet, and GoogLeNet, as well as more efficient Vision Transformer variants. Comparative analyses among these models may provide further insights into their suitability for remote sensing image classification. Furthermore, hybrid architectures integrating feature representations extracted from CNN and Vision Transformer models may be developed, where the combined feature vectors are classified using a unified SoftMax layer to improve classification accuracy. Nevertheless, such hybrid approaches are expected to require additional computational resources and training time.